# Question-focused Summarization by Decomposing Articles into Facts and Opinions and Retrieving Entities

**Krutika Sarode, Shashidhar Reddy Javaji,** and **Vishal Kalakonnavar**
College of Information and Computer Sciences
University of Massachusetts Amherst

## Abstract

This research focuses on utilizing natural language processing techniques to predict stock price fluctuations, with a specific interest in early detection of economic, political, social, and technological changes that can be leveraged for capturing market opportunities. The proposed approach includes the identification of salient facts and events from news articles, then use these facts to form tuples with entities which can be used to get summaries of market changes for particular entity and then finally combining all the summaries to form a final abstract summary of the whole article. The research aims to establish relationships between companies and entities through the analysis of Wikipedia data and articles from the Economist. Large Language Model GPT 3.5 is used for getting the summaries and also forming the final summary. The ultimate goal of this research is to develop a comprehensive system that can provide financial analysts and investors with more informed decision-making tools by enabling early detection of market trends and events.

## 1 Introduction

### 1.1 Task Description

We propose the development of a system that caters to financial analysts and investors by extracting relevant information from news articles and generating summaries on the potential impact on a particular company. This system comprises four key tasks. Firstly, it involves the extraction of sentences from news articles, which will be labeled as either factual or opinionated (F v O), with the focus on utilizing factual sentences. Since the news data (coming from the Economist) is generally accurate in its reporting, we constructed our baseline of F v O on linguistic structures. We classify subjective/opinionated sentences as containing subjective words/phrases ("I think", "believe", "hope", etc) and factual sentences as containing words/phrases like "according to", "studies show", and "research indicates". Secondly, the system will match each fact with the top-K most relevant entities. This is done using Pyserini to retrieve and match companies/their descriptions to an extracted fact. Thirdly, for each (fact, company) pair, the system will generate a summary outlining how the news will impact the particular company using a LLM/summarizer like GPT. We have currently found success using GPT 3/3.5 turbo. Finally, the system will aggregate the summaries across the entire article to extract the most significant information for each article, also using a LLM/summarize. GPT 3.5 turbo is currently being used for this task. By performing these tasks, the proposed system aims to provide financial analysts and investors with a comprehensive understanding of the potential effects of news articles on specific companies, allowing for more informed decision-making.

### 1.2 Motivation and Limitations of Existing Work

There exist works that exhibit proximity to our own work, although their approaches are distinct. Our work, however, leverages the potency of Large Language Models in tandem with facts and events extracted from news articles to yield outputs that are more efficacious. Prior works have been constrained by the utilization of knowledge graphs and other custom algorithms to establish relationships between given events or facts and the relevant news articles, thereby presenting certain limitations or inadequacies

### 1.3 Proposed Approach

The proposed methodology integrates a rule-based model and Large Language Models (LLM) for the extraction of facts and events from news articles. Using the extracted facts, a retrieval model is em-ployed to identify the top-K companies related to each fact. A tuple of (fact, company) is then generated for each relevant company. The impact of each fact on its respective company is determined by employing a language model. These individual summaries for each company are then aggregated to provide a concise overview of the potential impact of the news. This comprehensive summary can be utilized by financial analysts and investors to make informed decisions. By utilizing this approach, we aim to improve the accuracy and efficiency of analyzing news articles to better inform financial decision-making.

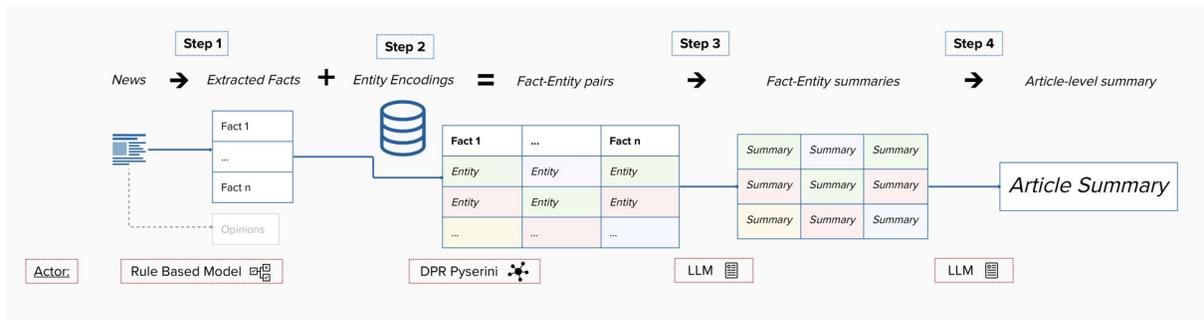

Figure 1: WorkFlow

1. **Factual news generation:** In order to utilize verified data that is based on evidence and avoid personal judgments and beliefs, it is necessary to separate factual sentences from news articles. This can be achieved by analyzing sentence structures and identifying subjectivity markers, such as words like "believe," "think," "feel," and "hope," to filter out opinionated sentences. Additionally, sentiment analysis and negation detection can be used to further refine the extraction of factual statements. This process involves using the NLTK and Spacy libraries in Python to create a rule-based model. By applying various vocabulary-based filters, we can identify and extract all factual statements, which can be used in subsequent steps.

2. **Generation of fact, entity pair:** After extracting facts from the news articles, the next step is to identify the three most relevant entities for each fact. In this context, relevance refers to the entities that are most likely to be impacted by the fact. To perform this step, we gather relevant information about the entities, such as their descriptions from Wikipedia pages, and use the Hugging Face sentence transformer to create embeddings for each entity. These embeddings are used to create an entity corpus. Next, we create an embedding for each fact and use Dense Passage Retrieval (DPR) techniques to map the fact embedding to the three most relevant entities in the corpus. DPR uses a bi-encoder to rank the entities based on their relevance to the fact. This process results in fact-entity pairs that provide a clear understanding of which entities are most likely to be impacted by each fact.

   **Example:**

   *Fact:*
   *"The legendary auto executive has spent the last two months in jail since his arrest in Tokyo on November 19. He has been charged by Japanese prosecutors with financial misconduct while head of automaker Nissan (NSANF). Ghosn, 64, denies the charges, but prosecutors have argued successfully that he should be kept in jail awaiting trial".*

   *Entity Description:*

   *Halliburton Company: "Halliburton Company is an American multinational corporation responsible for most of the world's hydraulic fracturing operations.[6] In 2009, it was the world's second largest oil field service company. It has operations in more than 70 countries.[7] It owns hundreds of subsidiaries, affiliates, branches, brands, and divisions worldwide and employs approximately 55,000 people.[8] The company has dual headquarters located in Houston and in Dubai and remains incorporated in the United States."*

3. **Generating impact of the fact on entity:** To determine the impact of each fact on the relevant entities, we utilize Language Learning

(LL) models such as the chat feature of the GPT-3.5 Turbo model. By inputting the factual statements and relevant entities, we can generate a brief description of how each entity is impacted by the fact. We use chat-based prompts to provide context for the LL models, such as: "Imagine that you are an expert financial analyst and you are analyzing the impact of a certain news article on different stocks and companies." To input the relevant entity, we use prompts like: "Consider the company " + entityname + ", which is described as " + entitydescription + "." Finally, we provide the factual statement to the LLM as part of the prompt and ask it to provide a brief summary of the potential impact of the news on the entity. This step helps us to gain a better understanding of how each fact is likely to affect the entities that are relevant to it.

**Example:**

*Fact-Entity Summary:*

*"The charge of financial misconduct against the former head of Nissan, who is also a former executive of Halliburton, may potentially affect the reputation of Halliburton negatively. As the company has been involved in controversies before, this news may lead investors and stakeholders to question the company's responsible management and ethical practices. This could potentially lead to a decrease in the company's stock price and investment, damaging its financial performance. However, the extent of the impact would depend on how the company and its current management address the issue and respond to the situation.".*

*Article summary:*

*"The news regarding Carlos Ghosn being replaced as chairman and CEO of Renault is not expected to have a direct impact on Alphabet Inc, General Motors Company (GM), Tesla Inc., Howmet Aerospace Inc., A garter, or Garman as they are not directly related to the automotive industry or the companies involved. However, there may be indirect effects on these companies and the broader technology and automotive sectors. For Alphabet Inc, any changes in the global automotive industry resulting from Ghosn's replacement could indirectly affect their self-driving car project, Waymo, and the overall market sentiment due to their global presence. Similarly, GM and Tesla Inc. may experience ripple effects within the automotive industry if the alliance between Renault and Nissan faces disruptions or changes, potentially impacting investor sentiment and the overall industry. Howmet Aerospace Inc. may be indirectly impacted if they have any partnerships or business agreements with Renault, Nissan, or Mitsubishi Motors, as mentioned in the news article. The extent of the impact would depend on the nature and scale of those collaborations. A garter and Garman are not directly associated with the automotive industry, so the news about Carlos Ghosn's arrest and detention is unlikely to have a significant impact on these companies. It is important for these companies to monitor the situation and potential ripple effects to determine any indirect impact on their operations and financial performance. However, without further information or developments, it is difficult to predict the exact impact on these companies"*

4. **Summary aggregation:** After generating individual summaries for each fact and entity, the next step is to combine them to create a comprehensive summary of the news article. To achieve this, we also make use of LL models and chat prompts to generate a final brief summary that covers all major points and arguments presented in the article. We provide a prompt that instructs the LL model to provide a concise summary that captures the most important information from the article. The summary should include key details such as the main topic, the impact on every relevant company, relevant facts and figures, and any important conclusions or recommendations presented in the text. We also ask the model to list every company that has been impacted by the news. By combining the individual summaries and generating a final comprehensive summary, we can provide a clear and concise overview of the news article, including its impact on relevant entities. This can be useful for decision-making and analysis purposes in various industries.

## 1.4 Likely Challenges and Mitigations

This task and methodology presented various challenges, such as obtaining outputs as expected from the language models, utilizing appropriate prompts for the outputs, finetuning models when necessary, and identifying the top-K companies related to a given fact. Should any of these challenges have arisen, a systematic approach was adopted to revise the plans. This involved a comprehensive review of the relevant literature to identify alternative methods and approaches. In case of unmitigated challenges, the problem statement was considered for revision to achieve feasibility with the existing approaches. Such an adaptive approach ensured the continued progress of the research while maintaining high standards of scientific rigor.

## 2 Related Work

[1] introduces example-based summarization techniques that circumvent the widely used query-based techniques that allow for more subjective and 'better'/desired results (in terms of the prompter's summarization goals). This improves drastically over query-based techniques by 1) providing more information to the system, 2) allowing the user to define their subjective prompt, 3) preventing the 'frustrating' process of iteratively improving a query to get the desired summary. The summarization step in our project aligns closely with this example based summarization technique in order to get a desirable summary of the aggregated facts from a news article by appending examples to input prompts.

[2] aims to tackle the challenges that appear in long form summarization: processing very long documents, nontrivial causal and temporal dependencies, and rich discourse structures[2]. Many widely used summarization techniques are used either for short form writing or long form writing in the form of scientific journals or patents: the latter of which does not contain a large possibility for openly interpretational summaries. BookSum tackles the long-form, creative-writing scene by creating an aggregate of summaries of creative works at a paragraph/chapters and full work level, it fine tunes baseline models to generate summaries at a paragraph level, ranks them using top-k, and combines the highly ranked summaries. This approach will differ from our approach as our data is based on financial news articles (short form text) and SEC data (long form straightforward data), so it would be better to rely on other summarization methodologies from other sources.

[3] aims to show how 'chain of thought' prompting can help improve a large language model's ability to perform complex tasks. By introducing an example QA pair, in which the answer contains an explanation the process/rationale (chain in thought) of solving the task, the LLM is able to imitate the process for a subsequent question to a higher degree of performance and accuracy than without the chain in thought addition. The idea of chain of thought prompting can be very helpful in our tasks in which we prompt a model to decide whether or not an event/entity has a positive or negative affect on another entity. By using this prompting technique, we can more easily see the effects that entities will have on each other, especially when looking at financial data.

[5] builds an idea known as 'self consistency', which builds off the success of a chain of thought prompting in order to replace the legacy decoding methodology of 'greedy decoding'. Self consistency is a self supervised approach of using multiple reasoning paths to come to an aggregated conclusive answer, like a group of humans would. This has shown to have a significant increase in commonsense reasoning/arithmetic benchmarks, to a rate as high as 17.9 percent (GSM8K). This process would be very useful again for prompting in both the effect of an event/entity on another entity (ex. 'chain-in-thought' QA pair appended to Q: "Does chatGPT have a negative/positive affect on Google") and 'summarization' tasks (the aggregation of facts from news sources) of our paper, as it would combine the success of chain in thought with the self-consistency route aggregation. Having multiple routes of thinking will be very helpful when considering financial effects of events/entities.

[6] aims to tackle the problem of fact verification in order to prevent misinformation by using a structured based approach of organizing text in tables as opposed to that of commonly used unstructured text. Each row on the table is structured with a verified claim, annotated evidence from wikipedia, and a label indicating whether or not the information is enough to reach a verdict/what the verdict is (support or refute the claim). Feverous' data structure can be a good way to model the factual or opinionated sentences from news text input. Also, its dataset could be used to finetune a LLM for fact extraction. One change we may have to make is to rely on more than wikipedia as a source of information, as it would be limited in terms of financial data but would be good for general company/entity

descriptions.

[7] aims to label whether or not an article is a news story (unbiased reporting) or an opinion article (persuasive reporting) in order to aid in fact checking. To create the dataset, the paper describes an approach of tagging sentences as either a 'claim' or a 'premise', with more 'premises' indicating a 'news story' and more 'claims' indicating an 'opinionated article'. After fine tuning various models with the data (BERT, SVM, RNN), they found that, during analysis, linguistic features that were highlighted in the dataset (negation, negation-suffix, digits and interjection, etc) did not affect the results, so it is a path to avoid taking when thinking of fact extraction. The paper also shows how we can fine tune the model to indicate whether text is opinionated or not, which can help in producing accurate results; however, the big difference between our task and the paper's is that we will be extracting factual and opinionated structures from text at a sentence level, while this paper focuses on an article level.

## 3 Experiments

### 3.1 Datasets

For our project, we have utilized a variety of news articles from the Economist dataset spanning from 2017 to 2019, along with additional data for later years through web scraping for news articles related to finance, business, world and economy. To create event encodings for companies, individuals, stocks, and commodities, we have obtained contextualized embeddings using data from Wikipedia. Furthermore, we are utilizing 10K sec filing data for creating company embeddings to incorporate business, risk and other key information of stakeholders.

### 3.2 Software

In the first step of extracting facts from a news article, we employ a rule-based technique using NLTK and Spacy libraries. These libraries assist in tokenization, Part-of-Speech tagging, as well as detecting negation and subjective markers. This aids in the extraction of factual sentences from the article. Subsequently, we utilize the Pyserini library for document retrieval and indexing. To incorporate dense vector representations (embeddings) for each predefined entity, we employ Dense Passage Retrieval techniques using Pyserini. For fact-summary creation, we utilize a chat-based model of LLM, such as gpt-3.5-turbo, along with various prompting techniques. The same approach is employed for abstract summary creation as well.

### 3.3 Challenges

One potential challenge we faced was with LLMs and creating good enough prompts to perform question-focused summarization using facts, zero-shot, and few-shot prompting. Prompting performance was a crucial step for the success of the project. Additionally, we lacked a standard metric for evaluating the abstract summaries. We found a lack of evaluation techniques that could determine the factual accuracy and usefulness of the generated summaries. An appropriate evaluation approach would involve comparing the generated summaries against factual truth, making the presence of domain experts with relevant knowledge crucial.

### 3.4 Contingency Plan

We aimed to attain zero-shot prompting for question-based summarization of the impact of news on different entities. If that did not work, we planned to use few-shot learning and feed more context to the prompt in order to generate better results. We experimented with different models to find the ones which gave the best possible response. While our prompt was able to generate results for the news impact on different companies, we aimed to generate better prompting techniques to generate better results and get more insights so that financial analysts can use this software in real-time in the future. In addition to that, we are devising plans to discover more effective evaluation techniques. This problem is highly open-ended and requires extensive experimentation to identify ways of assessing the quality of generated summaries. One approach is to create control and experiment sets and invite domain experts to evaluate the summaries based on their insightfulness. We can also investigate the impact of changing prompts on the evaluation scores. Additionally, we intend to employ ranking-based methods by treating human rankings as a baseline for relevance and calculating NDCG or Precision@n scores for the rankings generated by the language models. Currently, we are conducting this evaluation on a limited set of articles, but we aim to explore its effectiveness on a larger scale as well.

# 4 Results and Evaluation

## 4.1 Results

**News article:**

Some of the biggest names in tech are calling for artificial intelligence labs to stop the training of the most powerful AI systems for at least six months, citing "profound risks to society and humanity." Elon Musk was among the dozens of tech leaders, professors and researchers who signed the letter, which was published by the Future of Life Institute. The wave of attention around ChatGPT late last year helped renew an arms race among tech companies to develop and deploy similar AI tools in their products. OpenAI, Microsoft and Google are at the forefront of this trend, but IBM, Amazon, Baidu and Tencent are working on similar technologies.

> **Companies impacted:**
> 
> - Akamai Technologies, Inc.
> - American International Group, Inc. (AIG)
> - Bio-Rad Laboratories, Inc.
> - Intuitive Surgical, Inc.
> - Netflix, Inc.
> - Tesla, Inc. (TESS-l or TEZ-l)

**Factual Summary:** The companies that may be impacted include OpenAI, Microsoft, Google, Leidos, AIG, Tesla, and Netflix. For example, if AI tools have biased responses or spread misinformation, it could lead to increased insurance claims for AIG. The news of calls to stop the training of the most powerful AI systems for at least six months could have some impact on Tesla, Inc. as the company heavily relies on artificial intelligence in the development of their electric vehicles and related products. Netflix may need to reconsider its current business model and future strategies if AI tools are found to have biases or spread misinformation. Intuitive Surgical, which develops and markets robotic products designed to improve clinical outcomes of patients through minimally invasive surgery, might face questions about the safety and reliability of the da Vinci Surgical System due to potential risks and concerns associated with AI technologies. Bio-Rad Laboratories, which uses advanced technology in its products, may not be directly impacted as it is not involved in the development and deployment of AI tools.

## 4.2 Proposed Evaluation Method

Since we only had unlabelled data at our disposal, we lacked ground truth summaries for comparison purposes. The generated summaries needed to be evaluated on their factual accuracy regarding the impact of the news on relevant entities. Unfortunately, there was no predefined evaluation metric available for this specific criterion. We attempted to examine the stock movements over a certain period after the news was released, but this approach proved unreliable due to the multitude of factors influencing stock prices. Therefore, we opted for a contextual approach to assess the quality of the summaries, focusing on their conciseness, factualness, and usefulness for financial analysts in making decisions and conducting analysis.

The proposed evaluation method entails the use of ChatGPT, a language model, to evaluate the performance of summary generation. The method involves feeding the final summary along with a prompt, which outlines the metrics used to assess model performance, to ChatGPT. As there are no labeled data to evaluate summary coherence, the method is inspired by ChatEval, a tool that uses a human-level intervention to evaluate summary coherence. ChatGPT simulates this intervention, generating scores for coherence based on the alignment of the summary with the given information.

To determine the final scores for model performance, the process involves using multiple inputs and outputs, computing an average score for each metric of interest. Currently, the metrics being used for this evaluation are accuracy, conciseness, fluency, and engagement, with the possibility of more metrics being experimented on based on the outcomes of these.

To provide ChatGPT with the input required for generating summary coherence scores, two approaches are used: few-shot learning and the instructive method using prompts. The model is being evaluated using both approaches.

For the evaluation process, a single summary will be provided for a news article, and the task will be to rate the summary on the coherence metric, with a score range of 1 to 5. Coherence refers to the quality of all sentences collectively and is aligned with the DUC quality question of structure and coherence. A high-quality summary should be well-structured, well-organized, and build logically from sentence to sentence to provide a coherent body of information about a topic.

The evaluation process involves three steps. Firstly, the news article is carefully read to identify the main topic and key points. Secondly, the summary is compared to the news article to determine if it covers the main topic and key points and if it presents them in a clear and logical order. Lastly, a score for coherence is assigned based on the Evaluation Criteria, using a scale of 1 to 5, where 1 represents the lowest score, and 5 represents the highest.

|  | Factful | Concise | Useful |
|---|---|---|---|
| Instruction based | 4.92 | 4.0 | 3.5 |
| Instruction and one-shot | 4.42 | 3.71 | 3.10 |

Table 1: G-Eval Evaluation

**Example Prompt:**

You will be given one summary written for a news article. Your task is to rate the summary on one metric. Please make sure you read and understand these instructions carefully. Please keep this document open while reviewing, and refer to it as needed.

**Evaluation Criteria:**

- Coherence (1-5) - the collective quality of all sentences. We align this dimension with the DUC quality question of structure and coherence whereby "the summary should be well-structured and well-organized. The summary should not just be a heap of related information, but should build from sentence to sentence to a coherent body of information about a topic."

**Evaluation Steps:**

- Read the news article carefully and identify the main topic and key points.

- Read the summary and compare it to the news article. Check if the summary covers the main topic and key points of the news article and if it presents them in a clear and logical order.

- Assign a score for coherence on a scale of 1 to 5, where 1 is the lowest and 5 is the highest based on the Evaluation Criteria.

**Few Shot Prompt**
**Input:**
The quick brown fox jumps over the lazy dog.
**Output:**
The fox jumps over the dog.
**Evaluation:**
Accuracy: 100% Conciseness: 100% Fluency: 100% Engagement: 100%
**Input:**
The quick brown fox jumps over the lazy dog. The dog barks at the fox. The fox runs away.
**Output:**
The fox jumps over the dog, which barks at it. The fox runs away.
**Evaluation:**
Accuracy: 100% Conciseness: 80% Fluency: 100% Engagement: 100%
**Input:**
The quick brown fox jumps over the lazy dog. The dog barks at the fox. The fox runs away. The dog chases the fox. The fox hides in a hole. The dog looks for the fox, but cannot find it. The dog gives up and goes home.
**Output:**
The fox jumps over the dog, which barks at it. The fox runs away, but the dog chases it. The fox hides in a hole, and the dog looks for it, but cannot find it. The dog gives up and goes home.

**Evaluation:** Accuracy: 100%
Conciseness: 70%
Fluency: 100%
Engagement: 100%

**Input:**
The news about the battle between the world's biggest economies over semiconductor resources could potentially impact Microchip Technology Inc. (....................................) restrictions imposed by these economies could impact their accessibility and affordability, ultimately affecting A semiconductor's revenue and profits.
**Output:**
To manage the impact of the conflict, such as diversifying its supply chain or investing in research and development to produce more innovative products (....................................) hospitality brands such as Marriott and Hilton, retailers like Macy's and J.C. Penney, and manufacturers such as Ford and General Motors.

**Evaluation:**

**Generated response:**

Accuracy: 90%

Conciseness: 60%

Fluency: 90%

Engagement: 80%

While the output captures the main points from the input text and provides some valuable insights and recommendations, it also includes some irrelevant information about the COVID-19 pandemic and companies unrelated to the semiconductor industry. The text is also less concise and engaging, with longer and more complex sentences. However, it maintains good fluency, with a clear and coherent writing style.

**Human Evaluations**

We evaluated 20 articles using human evaluation and were asked to give a rating of 0-5 for the usefulness of the article aspect from the financial analyst perspective. The results we have received are decent and all of the article's summaries came out to be relevant with all of the articles being scored above 3. In addition to evaluating the summaries using ChatGPT, we also employed a ranking-based evaluation approach to assess Chat-eval's reliability. Since we needed to verify the accuracy of the results generated by ChatGPT, we devised a method involving rankings provided by a Computer Science PhD student for a small set of 5-10 articles. These rankings were based on the perceived usefulness of the summaries. This served as our baseline for comparison. We then obtained rankings from ChatGPT for the same articles and calculated NDCG and Precision@n scores to compare the two rankings. However, due to the limited number of articles used in this evaluation and the need for further scalability, we acknowledge that more extensive testing is necessary to validate the approach. This is an aspect we plan to explore in future research.

|  | Article 1 | Article 2 | Article 3 | Article 4 | Article 5 | Article 6 |
|---|---|---|---|---|---|---|
| Human ranking | 1 | 2 | 3 | 4 | 5 | 6 |
| LLM ranking | 2 | 1 | 3 | 5 | 4 | 6 |

Table 2: Human evaluation vs G-Eval

## 5 Future Steps

We aim to experiment with improved prompting techniques to produce more concise and precise results. To address this, we will explore sequential prompting techniques using Langchain to avoid repetition. Our plan involves expanding the entities included in our document store. Currently, we only have S&P 500 companies, but we intend to incorporate commodities such as gold and oil. Additionally, we aim to enhance the process of creating embeddings for these entities by utilizing their SEC filing data instead of relying solely on Wikipedia data. This shift is expected to provide more relevant information about the company and cover a broader range of aspects, ultimately improving the fact-entity mapping process and influencing subsequent steps in a positive manner. Our next step entails running the pipeline on a larger dataset to generate more robust evaluations. Furthermore, we will dedicate more effort to researching and identifying better evaluation techniques, with a particular focus on incorporating human evaluation methods. This approach will help in producing more insightful results.

# Appendix

## Experiments and Outputs

News article: WASHINGTON, DC AND MOSCOW P ATRIOT PARK in Kubinka, 60km south-west of Moscow, is a military Disneyland. Families can picnic among rows of Soviet-era aircraft. Children can frolic over tanks. Those doing so on January 23rd might have noticed a long green tube, studded with ridges and dials, roped off and watched by stern guards. This was not an exhibit. It was, supposedly, the canister for the 9M729 missile. Its launcher, an imposing truck, stood nearby, as Lieutenant- General Mikhail Matveyevsky, Russia's missile chief, pointed to a diagram of the missile's innards. "All tests of surface-to-surface missiles," he declared, "were conducted to a range that is less than the INF [Intermediate- Range Nuclear Forces] treaty limit." The show-and-tell did not impress America, whose diplomats had turned down an invitation to the theme park. On February 1st America declared itwould pull out of the INF treaty. It is exasperated not only with ten years of Russian cheating but also with the rapid growth in China's unshackled arsenal of over 2,000 missiles, 95% of which are of the range forbidden to America. "If Russia's doing it and if China's doing it, and we're adhering to the agreement," complained Donald Trump in October, "that's unacceptable". The pact will die once America's six months' notice expires in the summer. "The likelihood of compromise is zero," says Adam Thomson, Britain's envoy to NATO until 2016. That brings over 30 years of arms control to a close. The INF treaty was forged in 1987 to defuse a missile race between America and the Soviet Union. Intermediate-range nukes were appealing because they could hit key targets while remaining a safe distance away from the front line, without resorting to intercontinental ballistic missiles ( ICBM s). Appealing, but dangerous: ICBM s took 30 minutes to reach their targets; mid-range missiles got there in under ten. "It was like holding a gun to our head," remarked Mikhail Gorbachev. He and Ronald Reagan agreed to scrap all such land- based missiles, conventional and nuclear. By the 2000s the treaty began to chafe Russia. Its decrepit armed forces could not afford modern warships, submarines and warplanes to carry plentiful missiles, whose utility America had demonstrated with bombing campaigns in the Middle East and the Balkans. To Russia's south and east, countries like Israel, Iran, China and Pakistan were accumulating land-based missiles. In 2005 Russia's defence minister proposed that the treaty should be junked. Not long after came Russia's first test of the 9M729 . Since 2016 four battalions, roughly 100 missiles, have been deployed to two bases east of the Ural mountains and near the Caspian sea. "The 9M729 is core to Russian military thinking in terms of what they need to fight a regional war," says Pranay Vaddi, who worked on the issue for the State Department until October. American officials may decry the cheating. But they surely sympathise with the impulse. In recent years Pentagon officials have fretted over a widening missile gap in the Pacific. "China has a massive advantage over us," says a former American army official. "It cannot be overstated how important it is that we can field precision-guided missiles, unlimited by range."The INF treaty does not prohibit putting intermediate-range missiles on ships, submarines and aircraft. But these are expensive (a modern destroyer costs $1.8bn), demand manpower and have other things to do. Hence the appeal of land-based missiles. "A mobile TEL requires a couple of drivers and operators," says the former official, referring to the transporter-erector-launcher trucks used to fire missiles. "It is virtually impossible for the enemy to find." In a review of American nuclear posture last year, the Trump administration said it would respond to Russia's violation of the INF treaty by building a nuclear-tipped sea-launched cruise missile (which would be INF -compliant) and reviewing "concepts and options" for a conventional land-based one (which would not be). But a deployable weapon is some way off. The US Army is already working on a Precision Strike Missile ( P r SM ) due in 2023. Its range could easily be extended beyond the current INF ceiling of 499km. But even twice that would not get from Warsaw to Moscow. A longer-legged option would be to tweak the sea-based Tomahawk to fire from land; that is what America did during the INF crisis in the 1980s to produce the 2,500km-range Gryphon. Let's do launch But Pacific geography is forbidding. Guam, the likeliest host for American missiles in Asia if Japan demurs, is over 3,000km away from Shanghai. An entirely new missile would be required. Hypersonic boost-glide missiles, which

skip off the atmosphere at great speed, might fit the bill. But ground-launched ones are years away. Democrats, who took control of the House in January, have taken a dim view of the swelling defence budget. They may query why the Pentagon cannot make do with air- and sea-launched systems already in the pipeline. Nor is it obvious where new missiles would be put in Europe. Though NATO strongly backed America on February 1st, declaring that "Russia will bear sole responsibility for the end of the treaty," its members will be less keen on welcoming missiles, even non-nuclear ones. A few allies, like Poland, which is trying to seduce Mr Trump into setting up a new tank base, would probably embrace American arms on their soil. But a deal with Poland struck over NATO's head would compound anxiety over America's commitment to the alliance. It might also be seen as destabilising. "Missiles deployed on the territory of newer NATO members could reach Russia's main command points in less than five minutes," notes Pavel Zolotarev of the Russian Academy of Sciences. That, he warns, would allow the "destruction of its critical infrastructure without the use of nuclear warheads." In a crisis, a jittery Russia would not necessarily make wise choices. Another option for NATO would be to rejig the American-led missile-defence shield in Poland and Romania so that it could cope with Russian cruise missiles like the 9M729, rather than just Iranian and North Korean ballistic ones. Yet that might provoke an even bigger fight in NATO. The death of the INF treaty marks "the end of the post-cold war era," says Bruno Tertrais of the Foundation for Strategic Research, a French think-tank. What is clear is that the Trump administration has pulled the plug without a clear sense of how to navigate the era to come. This article was downloaded by calibre from https://www.economist.com/united-states/2019/02/09/america-calls-time-on-the-inf-treaty-ushering-in-a-new-age-of- missiles .

-*******************************

Facts: Patriot Park in Kubinka, Russia is a military-themed park with Soviet-era aircraft and tanks. Lieutenant-General Mikhail Matveyevsky, Russia's missile chief, displayed a diagram of the 9M729 missile's innards at Patriot Park. America declared on February 1st that it would pull out of the INF treaty. The INF treaty was created in 1987 to defuse a missile race between America and the Soviet Union. The treaty prohibited intermediate-range nukes, which could hit key targets while remaining a safe distance away from the front line. Russia began to chafe under the treaty in the 2000s, as it could not afford modern warships, submarines, and warplanes to carry plentiful missiles. Since 2016, Russia has deployed roughly 100 9M729 missiles to two bases east of the Ural mountains and near the Caspian sea. America is exasperated with ten years of Russian cheating and the rapid growth in China's unshackled arsenal of over 2,000 missiles, 95The Trump administration plans to respond to Russia's violation of the INF treaty by building a nuclear-tipped sea-launched cruise missile and reviewing "concepts and options" for a conventional land-based one.. Democrats, who took control of the House in January, may query why the Pentagon cannot make do with air- and sea-launched systems already in the pipeline. . NATO strongly backed America on February 1st, declaring that "Russia will bear sole responsibility for the end of the treaty.". Pavel Zolotarev of the Russian Academy of Sciences warns that missiles deployed on the territory of newer NATO members could reach Russia's main command points in less than five minutes, allowing the "destruction of its critical infrastructure without the use of nuclear warheads." . The death of the INF treaty marks "the end of the post-cold war era," according to Bruno Tertrais of the Foundation for Strategic Research.

Companies impacted: ['Intercontinental Exchange, Inc. (ICE) ', 'F5, Inc. ', 'American International Group, Inc. (AIG) ', 'Leidos, formerly known as Science Applications International Corporation (SAIC), ', 'Trimble Inc. ', 'Vulcan Materials Company (NYSE: VMC) ', 'The Interpublic Group of Companies, Inc. (IPG) ', 'The Lockheed Martin Corporation ', 'IQVIA, formerly Quintiles and IMS Health, Inc., ', 'Howmet Aerospace Inc. (formerly Arconic Inc.) ', 'American Airlines Group Inc. ', 'The Union Pacific Corporation ', 'L3Harr', 'Align Technology ', 'Gen Digital Inc. (formerly Symantec Corporation and NortonLifeLock) ', 'Raytheon Technologies Corporation ', 'Halliburton Company ', 'Broadcom Inc. ', 'Invesco Ltd. ', 'Northrop Grumman Corporation '].
-

Overall summary: The news of America withdrawing from the INF treaty and the subsequent increase in arms race may have a positive impact on Northrop Grumman Corporation. As a major defense technology provider, the increased demand for advanced missile systems and defense equipment could lead to an increase in revenue and growth opportunities for the company. However, the potential negative impact of political tensions and instability in the global arms race should also be considered. The news about the end of the INF treaty, which prohibited land-based intermediate-range missiles, could potentially impact Howmet Aerospace Inc. due to its manufacturing of components for jet engines and titanium structures for aerospace applications. The increase in missile production and development could lead to a potential increase in defense spending and a rise in demand for Howmet's products. The US decision to pull out of the INF treaty could potentially have a positive impact on The Lockheed Martin Corporation. As one of the largest defense contractors in the world, it may receive further contracts and revenue from the US government to build and develop land-based missiles to compete with Russia and China's growing missile arsenal. The article discusses the end of the INF treaty and the potential for a new arms race, particularly in the development and deployment of intermediate-range missiles. As Leidos provides technical services to the United States Department of Defense and other government agencies, particularly in the fields of defense and aviation, it may benefit from increased demand for its expertise and services as the US seeks to modernize and expand its missile capabilities. However, there may also be increased competition and pressure on Leidos to provide cutting-edge solutions in this area. The news about America's withdrawal from the INF treaty and the possibility of an arms race may impact the construction industry as a whole, but it is unlikely to have a significant direct impact on Vulcan Materials Company (NYSE: VMC) as they primarily produce and distribute construction materials. However, the potential escalation of tensions may impact the construction industry's demand for their products, depending on how the situation develops. The news about the US pulling out of the INF treaty and the potential for a new arms race between America, Russia, and China is not expected to have a significant impact on American International Group, Inc. (AIG), as the company operates primarily in the insurance and financial sectors. However, any geopolitical instability or conflict resulting from this situation could affect the global markets and potentially impact AIG's investments and operations. The news about the INF treaty and Russia's missile program is unlikely to have a direct impact on Align Technology. The company does not manufacture any military equipment or have any significant operations in Russia. However, the broader geopolitical implications of this news could impact the global economy and potentially affect the demand for Align's products in certain regions. The news article does not have a direct impact on The Union Pacific Corporation since it is a railroad holding company. However, it may indirectly impact the transportation industry as tensions rise between the United States and Russia, and potentially China. Any increase in military spending or conflict may divert resources away from infrastructure and transportation projects, affecting Union Pacific's business. The news about America pulling out of the INF treaty, which led to discussions about the missile race and the development of new weapons, is unlikely to have a significant impact on Invesco Ltd, an investment management company based in the US. The US government's decision to pull out of the INF treaty, which could lead to a new arms race, may result in increased demand for Raytheon's missile defense systems, as well as opportunities for the company's missile division to develop and supply new land-based missiles. However, there may also be increasing competition from other defense manufacturers in the industry. The news about America pulling out of the INF treaty and the potential increase in missile race could lead to increased demand for semiconductor and infrastructure software products produced by Broadcom Inc. as countries like America and China may escalate their military capabilities in this area. The news article about America pulling out of the INF treaty and the potential growth in missile technology could indirectly impact Trimble Inc, as they are involved in the development of satellite navigation systems and other hardware technologies that could be used in defense and military applications. The potential increase in demand for precision-guided missiles could lead to an increase in demand for Trimble's technology. The news article does not directly relate to Gen Digital Inc. or the cybersecurity industry, so it is unlikely to have a significant impact on the company's stock or operations. However, any potential global instability resulting from the end of the INF treaty and the escalation of military tension between the US and Russia could potentially affect the overall market and investor sentiment, including towards

Gen Digital Inc. The news about America pulling out of the INF treaty and the potential for a missile race could have a positive impact on L3Harr, as the company specializes in defense and surveillance solutions. The company may see increased demand for their products and services due to the growing concern over missile capabilities of the US and its rivals. The news article about America's withdrawal from the INF treaty and the potential development of new missiles could have an indirect impact on Halliburton Company, as it is a major player in the energy services sector, which includes the production and transportation of oil and gas. Any increased tensions or conflicts resulting from the development and deployment of new missiles could potentially disrupt the industry, leading to changes in demand for Halliburton's services and products. The news about the potential global arms race and increasing military tension between the US, Russia, and China is unlikely to have a direct impact on F5, Inc. as a technology company. However, if the situation escalates into a full-blown conflict, it could have a negative impact on the overall global economy, which in turn may lead to a decrease in demand for F5's products and services. The news about the US pulling out of the INF treaty and the potential for a new missile race could potentially impact American Airlines Group Inc. due to the increased geopolitical tensions and potential for military action, leading to a decrease in air travel demand and revenue for the company. The news about the US pulling out of the INF treaty and the potential rise of missile development could potentially lead to an increase in military spending and a shift in focus for the US government. However, it is unlikely to have a direct impact on The Interpublic Group of Companies, Inc. (IPG), which operates in the advertising industry. The news article about America pulling out of the INF treaty and the potential missile race to come would not have a direct impact on Intercontinental Exchange, Inc. (ICE), which mainly operates global financial exchanges and clearing houses. However, it may indirectly impact the financial markets and commodities that ICE exchanges deal with due to the geopolitical tensions that may arise from such developments. Based on the news article, it is unlikely that the INF treaty withdrawal will directly impact IQVIA. IQVIA primarily operates within the biopharmaceutical industry and is not directly involved in military or defense matters. However, any potential escalation of tensions between the US and Russia or China could lead to broader economic and geopolitical impacts that may indirectly affect IQVIA's business..
-*************************************

Final abstract instruction based 1:

The US withdrawal from the INF treaty and the potential for an arms race may have a positive impact on defense technology providers such as Northrop Grumman and Lockheed Martin, as well as Howmet Aerospace due to increased demand for missile systems and defense equipment. Leidos, which provides technical services to the US Department of Defense, may also benefit from increased demand for its expertise. The news is unlikely to have a significant direct impact on Vulcan Materials Company and American International Group, but any geopolitical instability resulting from the situation could affect global markets and potentially impact AIG's investments and operations. Align Technology is not expected to be directly impacted by the news.

The news of America pulling out of the INF treaty could impact the global economy and potentially affect the demand for Align's products in certain regions. The transportation industry, including Union Pacific Corporation, may be indirectly impacted as tensions rise between the US and Russia. Raytheon's missile defense systems may see increased demand due to the potential increase in the missile race, while Broadcom Inc. and Trimble Inc. may also see increased demand for their products. Gen Digital Inc. may not be directly impacted, but any potential global instability resulting from the end of the INF treaty could affect the overall market and investor sentiment. L3Harr, specializing in defense and surveillance solutions, may see increased demand for their products and services. Halliburton Company, a major player in the energy services sector, may also be indirectly impacted by any increased tensions or conflicts resulting from the development.

The potential global arms race and increasing military tension between the US, Russia, and China could impact Halliburton's services and products. F5, Inc. may be indirectly impacted if the situation escalates into a full-blown conflict, leading to a decrease in demand for its products and services. American Airlines Group Inc. may be impacted due to the increased geopolitical tensions and potential for military action, leading to a decrease in air travel demand and revenue. The Interpublic Group of Companies, Inc. (IPG) and Intercontinental Exchange, Inc. (ICE) are unlikely to be directly impacted, but may be indirectly impacted by the geopolitical tensions that may arise from such developments. IQVIA is not directly involved in military or defense matters, but any potential escalation of tensions between the US and Russia or China could lead to broader economic and geopolitical impacts that may indirectly affect its business..
-\*\*\*\*\*\*\*\*\*\*\*\*\*\*\*\*\*\*\*\*\*\*\*\*\*\*\*\*\*\*\*\*\*\*\*\*\*\*\*\*

Final abstract instruction and one shot 2:

The news of America withdrawing from the INF treaty and the potential increase in the arms race may have a positive impact on defense contractors such as Northrop Grumman Corporation, The Lockheed Martin Corporation, and Leidos. Howmet Aerospace Inc. may also benefit from increased demand for its products due to the potential increase in missile production and development. The news is unlikely to have a significant impact on Vulcan Materials Company and American International Group, Inc. However, any geopolitical instability resulting from this situation could affect the global markets and potentially impact AIG's investments and operations. The news is not expected to have a direct impact on Align Technology.

The news of America pulling out of the INF treaty and the potential for a missile race could have a positive impact on defense contractors, such as Raytheon and L3Harris, as well as companies involved in semiconductor and infrastructure software products, such as Broadcom. Trimble, which develops satellite navigation systems, may also see increased demand for their technology. The news may indirectly impact transportation companies, such as Union Pacific, and energy services companies, such as Halliburton, due to potential resource diversion and global instability. The news is unlikely to have a significant impact on Invesco or Gen Digital.

The potential global arms race and increasing military tension between the US, Russia, and China could impact defense contractors such as Halliburton, but is unlikely to have a direct impact on technology companies like F5, Inc. However, if the situation escalates into a full-blown conflict, it could negatively impact the overall global economy, leading to a decrease in demand for F5's products and services. The news about the US pulling out of the INF treaty and the potential for a new missile race could potentially impact American Airlines Group Inc. due to the increased geopolitical tensions and potential for military action, leading to a decrease in air travel demand and revenue for the company. The news may indirectly impact the financial markets and commodities that Intercontinental Exchange, Inc. (ICE) exchanges deal with due to the geopolitical tensions that may arise from such developments. The INF treaty withdrawal is unlikely to directly impact IQVIA, but any potential escalation of tensions between the US and Russia or China could lead to broader economic and geopolitical impacts that may indirectly affect IQVIA's business..
-\*\*\*\*\*\*\*\*\*\*\*\*\*\*\*\*\*\*\*\*\*\*\*\*\*\*\*\*\*\*\*\*\*\*\*\*\*\*\*\*

Instruction based abstract eval scores 1:
Factfulness: 5
Conciseness: 3
Usefulness: 4

The summary provides relevant and accurate information on the impact of America's withdrawal from the INF treaty. It specifically highlights companies that could benefit from increased demand for defense-related products and services, such as Northrop Grumman, Lockheed Martin, and Leidos. However, the summary also notes that the situation may indirectly impact other companies in the global market due to potential geopolitical instability. The summary could be more concise in its discussion of how the news may impact other companies, but overall it provides helpful information to a financial analyst..
-\*\*\*\*\*\*\*\*\*\*\*\*\*\*\*\*\*\*\*\*\*\*\*\*\*\*\*\*\*\*\*\*\*\*\*\*\*\*\*\*

Instruction based abstract eval scores 2:
News 1:

Factfulness: 5 The summary provides accurate and relevant information about the impact of the news article on the listed companies. The writer discusses specific ways in which the INF treaty withdrawal may impact various companies.

Conciseness: 4 The summary provides essential information in a concise manner, but some statements could be more straightforward and less repetitive.

Usefulness: 4 The summary provides useful data to a financial analyst to analyze the market and identify potential impacts on specific companies. However, the potential impact on other industries or markets is not discussed, which limits the overall usefulness of the summary.

Overall, the summary accurately and concisely summarizes the impact of the news article on the listed companies, but could provide more comprehensive analysis of potential impacts on other industries or markets.

News 2:

Factfulness: 5 The summary provides accurate and relevant information about the potential impact of the news article on the listed companies. The writer discusses specific ways in which the increased global arms race may impact various companies.

Conciseness: 4 The summary provides essential information in a concise manner, but some statements could be more straightforward and less repetitive.

Usefulness: 4 The summary provides useful data to a financial analyst to analyze the market and identify potential impacts on specific companies. However, the potential impact on other industries or markets is not discussed, which limits the overall usefulness of the summary.

Overall, the summary accurately and concisely summarizes the potential impact of the news article on the listed companies, but could provide more comprehensive analysis of potential impacts on other industries or markets.. -**************************************

Instruction and one shot based abstract eval desc 1:

Coherence: 5 - The summary is well-structured and each sentence logically follows the previous one, providing a clear flow of information.

Accuracy: 4 - The summary accurately explains how the US withdrawal from the INF treaty could impact defense technology providers such as Northrop Grumman and Lockheed Martin, although the impact on other companies listed is more uncertain and indirect. The summary also correctly identifies the potential for geopolitical instability and its impact on global markets, and the potential demand for Raytheon's missile defence systems. However, the summary's assessment of the impact on other companies, such as F5, Inc. and American Airlines Group Inc., is more speculative and may not necessarily come true.. -**************************************

Instruction and one shot based abstract eval desc 2:

Coherence: 4 The summary is mostly well-structured and organized, with clear transitions between sentences. However, some sentences could be improved for clarity and to better connect to the main topic.

Accuracy: 3 The summary mentions some potential impacts on defense contractors, such as Northrop Grumman Corporation and The Lockheed Martin Corporation, but does not provide a clear explanation of why they would be affected. The potential impact on transportation companies and energy services companies, as well as the indirect impact on financial markets and commodities, is mentioned, but not fully explained. The summary does not address the specific companies listed in the prompt or provide a clear explanation of how the news article may impact them. Overall, the summary includes some important points about potential economic impacts of the INF treaty withdrawal, but could benefit from more specific examples and explanations..